\documentclass{article}

\usepackage[final, nonatbib]{ARLET_2024}

\usepackage{graphicx}

\usepackage[utf8]{inputenc} % allow utf-8 input
\usepackage[T1]{fontenc}    % use 8-bit T1 fonts
\usepackage{hyperref}       % hyperlinks
\usepackage{url}            % simple URL typesetting
\usepackage{booktabs}       % professional-quality tables
\usepackage{amsfonts}       % blackboard math symbols
\usepackage{nicefrac}       % compact symbols for 1/2, etc.
\usepackage{microtype}      % microtypography
\usepackage{xcolor}         % colors
\usepackage{authblk}        % multiple affiliations
\usepackage{biblatex}       % Imports biblatex package
\title{Open Problem: Active Representation Learning}

\author[1,2]{Nikola Milosevic}
\author[1,3]{Gesine Müller}
\author[3]{Jan Huisken}
\author[1,2]{Nico Scherf}

\affil[1]{Max Planck Institute for Human Cognitive and Brain Sciences, Leipzig 
}
\affil[2]{Center for Scalable Data Analytics and Artificial Intelligence (ScaDS.AI), Dresden/Leipzig}
\affil[3]{Multiscale Biology, Georg-August-Universität Göttingen, Göttingen
\texttt{\{nmilosevic,nscherf\}@cbs.mpg.de}
\texttt{\{gesinefiona.mueller,jan.huisken\}@uni-goettingen.de}}

\begin{document}

\maketitle

\begin{abstract}
  In this work, we introduce Active Representation Learning, a class of problems that intertwines exploration and representation learning within partially observable environments. 
  We extend ideas from Active Simultaneous Localization and Mapping (active SLAM), and translate them to scientific discovery problems, exemplified by adaptive microscopy. 
  We explore the need for a framework that derives exploration skills from representations, aiming to enhance the efficiency and effectiveness of data collection and model building in the natural sciences. 
\end{abstract}

\section{Introduction}

Representation learning is frequently used to uncover the latent structure in a data set and lies at the core of modern data science and machine learning. 
Deep learning-driven representation learning approaches have been widely applied in cases where data is abundant but challenging to interpret, for example in scientific data analysis.
Frequently, such datasets cannot be labeled by human experts, because the nature of the scientific questions does not allow for clear labeling.
For example, recent methods in the analysis of functional neuroimaging data rely on self-supervised learning techniques to uncover the geometric properties of the data set, often using methods like auto-encoding~\cite{Pandarinath2018-dp}, or contrastive learning~\cite{Schneider2023-rn}. 
While these approaches can lead to remarkably insightful representations of complex data structures, they are mostly applied to scenarios where large, well-structured datasets are readily available for training general-purpose deep learning models.

However, in some data science problems, data has to be acquired by \textit{exploring} a complex physical system whilst interacting with it through measurements.
Consider, for example, active sensing-type problems, where data is collected through probing the input-output behavior of a black box process, including medical diagnosis, exploratory data analysis, active simultaneous localization and mapping (active SLAM), and adaptive light-sheet microscopy.
Especially in light-sheet microscopy, we may find ourselves with the potential to collect, through a costly process, vast amounts of data of which only a small subset is truly informative about the questions and tasks at hand.

Knowing how and where to collect new data is challenging, but it can make representation learning more effective. In turn, having a representation of the underlying dynamical system and the measurement mechanism, potentially helps guiding data acquisition.
However, such an active approach to representation learning requires integrating aspects of sequential decision making with representation learning concepts.
This raises the question: 
\textit{How can we integrate representation learning and sequential decision making?}

\begin{figure}[ht]
    \centering
    \includegraphics[width=\textwidth]{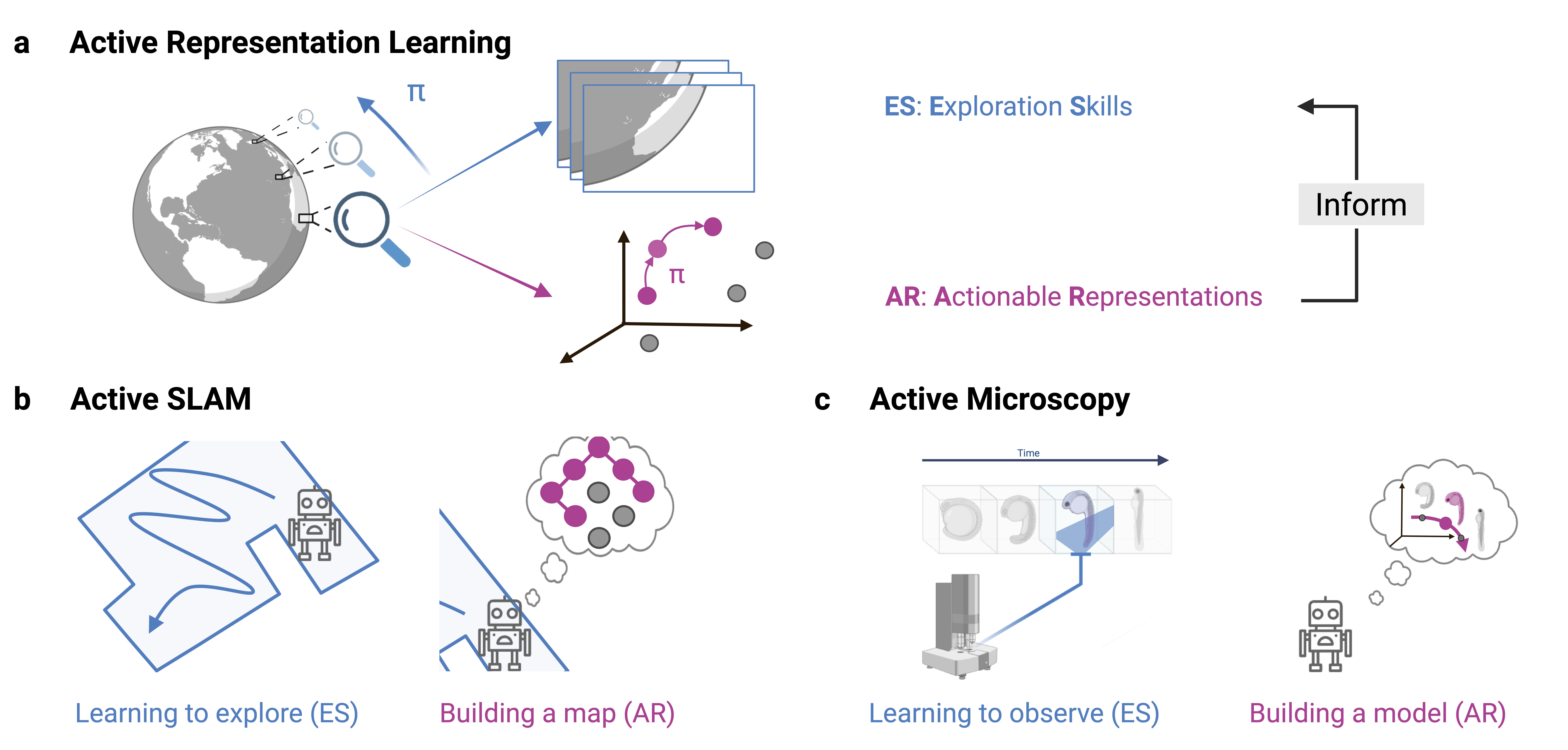}
    \caption{Active representation learning (ARL) is a set of problems that involves deriving Exploration Skills (ES), informed by Actionable Representations (AR), to learn a suitable exploration policy $\pi$, see  Figure a. Two common examples are active SLAM (Figure b) and active microscopy (Figure c).}
    \label{fig:pull-figure}
\end{figure}

We observe that similar questions appear in a variety of problem domains, which motivates a larger class of problems we call \emph{Active Representation Learning}, sketched out in Figure~\ref{fig:pull-figure}.

\section{Related Work}

\paragraph{From Adaptive to Active Microscopy.}
ARL is inspired by the challenge of building an \textit{active microscope}.
With the invention of Selective Plane Illumination Microscopy (SPIM;~\cite{huisken_2004}) the field of microscopy has entered the realm of real-time high data acquisition of biological processes, that opens the door for adaptive microscopy~\cite{Scherf2015-ji, Daetwyler2023-zu}.
The terms \textit{adaptive} or \textit{smart} microscopy are used for a wide range of solutions to common problems in microscopy and encompasses methods with the goal of maximizing the information content of the imaged volume. 
They aim for providing assistance throughout the imaging process, which includes optimizing the optical system during an imaging experiment ~\cite{Royer2016-dz, quirin_2016, McDole2018-ek} , managing and representing data \cite{cheeseman_2018}, applying processing techniques 
to improve image quality \cite{guo_2020}, detecting events of interest \cite{mahecic_2022, alvelid_2022}, and incorporating prior knowledge \cite{he_2020, gritti_2022}. 
We here want to further coin the term \textit{active microscopy} where an underlying model of the whole dynamical system including the biological process under the microscope and the imaging mechanics is exploited to make active decisions that lead to high-quality observations of the sample. 

\paragraph{Representations in Online Learning}\label{sec:related}
Representation learning and online learning can come together in different ways, two of which are particularly relevant to ARL.
One problem formulation is to allow prior knowledge to improve sample efficiency, either in the form of given, pre-trained or shared representations~\cite{agarwal2023provablebenefitsrepresentationaltransfer, cella2023multitaskrepresentationlearningstochastic, yang2021impactrepresentationlearninglinear,zhang2022makinglinearmdpspractical} or through optimistic approaches to online learning~\cite{rakhlin2014onlinelearningpredictablesequences, flaspohler2021onlinelearningoptimismdelay}.
While the theory behind these approaches is relevant to ARL, we want to learn to make decisions about how to best collect data for \textit{learning} the representations in the first place.
Another stream of work considers practical aspects of representation learning in continual settings, either with~\cite{hayes2019memoryefficientexperiencereplay, hayes2020remindneuralnetworkprevent} or without~\cite{rao2019continualunsupervisedrepresentationlearning, ren2022onlineunsupervisedlearningvisual} supervision. Continual unsupervised representation learning is an especially important aspects of ARL. 
However, it is not enough to address ARL, since it usually doesn't consider the agent's decision in acquiring new data for representation learning.

\paragraph{Representations in Deep Reinforcement Learning.}

Learning coherent state representations has been argued to be beneficial for deriving directed exploration strategies~\cite{yarats2021reinforcement}.
In problems where the state space has a known structure, the representation learning problem can be simplified through inductive biases specific to the task and integrated with exploration strategies, as in visual active SLAM using Deep Reinforcement Learning \cite{Chaplot_undated-lz, Chaplot2020-fn, Chaplot2020-vo}.
Learning representations using world models \cite{hafner2019learning, hafner2019dream}, contrastive learning \cite{laskin2020curl,stooke2021decoupling, yarats2021reinforcement}, or data augmentation~\cite{laskin2020reinforcement} has demonstrated remarkable success in improving sample efficiency in reinforcement learning. 
Perhaps strikingly, such representations have even been shown to drive behavior in the absence of a specific task or reward signal~\cite{laskin2021urlb}.
ARL is motivated by exactly this kind of representation-driven decision making.
While the works mentioned by \cite{laskin2021urlb} 
demonstrate successful proofs of principle, they do not address the continual learning aspects mentioned earlier or partial observability, and their properties and limits are yet too poorly understood to be usable in real-world practical settings.

\section{Active Representation Learning}

\subsection{Motivation}
Active Representation Learning is motivated by the concept of the active microscope. In developmental biology, the goal is often to uncover the 3D structure and dynamical processes of a biological specimens' developmental trajectory. 
To this end, one tries to acquire a stack of images in the fastest and least invasive way possible.
It might be tempting to blindly acquire as many planes as possible, however, imaging a biological process (e.\,g.\ a developing organism) is inherently subjected to the trade-off involving spatial and temporal resolution, signal-to-noise ratio and sample health. Additionally, absorption, scattering, and refraction lead to incomplete and noisy 2D image planes.

Modern light-sheet microscopes can quickly generate enormous amounts of data in the range of multiple terabytes, thus predicting which measurements are actually necessary would considerably simplify the data processing pipeline or even makes it feasible in the first place. 
Because we are seeking a concise representation of the biological process (see e.\ g.\ ~\cite{Kobayashi2022-jh}), the raw data is just a means to an end. An \textit{active} microscope would be able to balance the imaging trade-offs, and learn a useful representation of the biological process while being as gentle with the specimen as possible~\cite{Scherf2015-ji}. This also involves learning a representation of the imaging process to inform the decision about \textit{how}, \textit{where} and \textit{when} to look next for new data. This can be achieved by separating the dynamical factors that are directly controllable (where and how to look), and those that are not (the biological process itself), learning to control what can be controlled with minimal cost, and motivating the agent to curiously explore what can not. 

Due to the cumulative nature of the costs incurred, and the temporal dependencies between the continual formation of representations and the agent's decisions about how to operate the microscope, this is a problem at the intersection of sequential decision making and continual representation learning.

\subsection{Relation to Other Decision Frameworks}
If our only problem were to search for a good set of microscope parameters to obtain high-quality views of the biological specimen, one would rightfully consider this a bandit task.
It is already known from previous works (see Sec. ~\ref{sec:related}), that external knowledge about the data generating process (e.g. in the form of representations) can be shown to improve sample complexity of decision algorithms. However, what sets the ARL problem apart is that state representations are not only unknown, learning them is the primary goal, and deriving a suitable learning objective is part of the problem.

During data acquisition, the potentially incomplete representation must be used as a basis for informing the next action to take, similar to how the incompleteness of the map in active SLAM is used to inform where to go next. 
This circular dependency of representation and decision making is the basis of ARL and its framing as a generalization of active SLAM.
In general, ARL is difficult to frame as a bandit, since the agent-environment system has a complex state.

\subsection{From Active SLAM to Active Microscopy}
The basic idea of Active Representation Learning (ARL) is that a data structure is in some sense embedded in a Partially Observable Markov Decision Process (POMDP;~\cite{Astrom1965-ym}, see Appendix~\ref{appendix:pomdp}).
ARL is inspired by active SLAM, where we interpret the map as an example of a data structure embedded in a POMDP. 
Following SLAM, we model ARL assuming a general belief-dependent utility instead of an external reward~\cite{Araya-Lopez_undated-qy}.

Active SLAM introduces further specifications to the POMDP formalism in the form of a structured state-space.
The state at time $t$ is usually assumed to be a combination of the robot's pose $\mathbf{x}_t \in \mathcal{X}$ and its internal representation (a map) $\mathbf{m}_t \in \mathcal{M}$ of the external state space. 
This results in a factored state space of the form $\mathcal{S}=\mathcal{X}\times\mathcal{M}$~\cite{Placed2022-mf}.
While, in active SLAM, $\mathbf{m}_t \in \mathcal{M}$ refers to a representation of 2D navigable space, we want it to refer instead to a more general notion of model that includes controllable and uncontrollable dynamic factors.
Active microscopy, for example, factors as $\mathcal{S}=\mathcal{X}\times\mathcal{M}^c\times\mathcal{M}^u$, consisting of \begin{itemize}
    \item the microscope's directly controllable parameters $\mathcal{X}$, e.\,g.\ the laser power used for excitation,
    \item indirectly controllable factors of the environment $\mathcal{M}^c$, including e.\,g.\ image quality,
    \item and uncontrollable factors $\mathcal{M}^u$, e.\,g.\ the latent biological processes we wish to model.
\end{itemize}

A setup like this is not only reminiscent of active SLAM, but also of reinforcement learning-based visual attention~\cite{minut2001}.
We can easily think of the microscope setup as a robotic environment in the context of RL. The controllable parts of the microscope act like a body, whereas the imaged volume is the outside environment. Our agent has to learn how to influence its observations, and how to do so in order to perceive the most interesting aspects of the environment's uncontrollable parts.

\subsection{Challenges}

An immediate difficulty is that there is no specific reward function. While we can choose from a wide variety of intrinsic motivation objectives from the literature, most are not stationary functions of state-action pairs, but depend on some internal belief state (see $\rho$-POMDP; \cite{Araya-Lopez_undated-qy}).
Deriving suitable utility functions is an open problem that has received much attention in Unsupervised RL recently, and is also a known problem in SLAM.
How the latent state space should be represented by the agent and how uncertainty about it relates to the agent's utility is another long-standing problem in active SLAM~\cite{Placed2022-mf}, and also in Reinforcement Learning, see e.g. \cite{Erraqabi2021-xi}.
The non-stationarity and partial observability of the resulting decision problem, as well as the continuity of representation learning present significant challenges for practical implementation.
Beyond practical considerations, theoretical questions abound, involving the identifiability~\cite{Hyvarinen2023-gn} of the controllable and uncontrollable latent processes, and properties of the exploration behaviors that result from various utility functions.
Finally, since we are usually dealing with physical dynamical systems, control theoretic issues can not be ignored, including controllability and reachability notions for the latent state-space, observability of the latent processes, and stability of exploration skills.

\subsection{Opportunities}

Active Representation Learning highlights the need for a framework capable of capturing model building through sequential, incomplete observations, and the need to learn the underlying structure of a latent object or phenomenon. Ideally, the intelligent agent can learn structural similarities across a larger class of similar problems and use this information the next time a similarly structured problem is encountered, e.g. when a similar specimen is examined under the microscope.
We believe that the overarching themes of Active Representation Learning transcend applications in robotics and microscopy.
A large interdisciplinary community would benefit from a formal treatment of intelligent model building systems. 
A common framework would allow scientists from different fields to come together and share their experiences with related problems.
Ideally, the community would work towards and eventually achieve a \textit{virtual scientist} framework that allows us to build agents that can curiously explore complex processes the way humans do, potentially discovering and exploiting its symmetries.

\section{Conclusion}
Active Representation Learning (ARL) presents a step forward in the intersection of decision making and representation learning within partially observable environments. 
By leveraging the concepts and techniques from active SLAM, we have illustrated how these ideas can be extended to a larger class of problems with a similar structure, exemplified by what we call active microscopy.
The proposed ARL framework emphasizes the importance of disentangling controllable and uncontrollable factors in the environment, thus enabling intelligent agents to make informed decisions about where and how to explore. 
We argue that this approach not only improves the quality of the data collected but also facilitates the creation of more robust and interpretable models of complex systems.
Future research should focus on refining the theoretical foundations of ARL, addressing practical implementation issues, and exploring new applications across various domains. 
Ultimately, ARL holds the promise of transforming how we interact with and understand complex, dynamic systems, paving the way for breakthroughs in both artificial intelligence and scientific discovery.

\section*{Acknowledgments}
The authors would like to thank Simon Hirländer for insightful comments on the practical issues in ARL, and the anonymous reviewer for valuable feedback on the connections of ARL with existing methods. N. M. and N.S. are supported by BMBF (Federal Ministry of Education and Research) through ACONITE (01IS22065) and the Center for Scalable Data Analytics and Artificial Intelligence (ScaDS.AI.) Leipzig. N.M. is also supported by the Max Planck IMPRS CoNI Doctoral Program.
J.H. is supported by the Alexander von Humboldt Foundation (Alexander von Humboldt Professorship; J.H.) and the German Research Foundation (Germany’s Excellence Strategy EXC 2067/1-390729940; J.H.).
G.M. is supported by the MWK (Niedersächsisches Ministerium für Wissenschaft und Kultur, 6707040) and is member of the Hertha Sponer College of the Cluster of Excellence Multiscale Bioimaging (MBExC; EXC 2067/1- 390729940), University of Goettingen, Germany.
Figure 1 was created using BioRender.com.

\printbibliography

\appendix
\section{Extended Background}

\subsection{Background on POMDPs and Active SLAM}\label{appendix:pomdp}
A Partially Observable Markov Decision Process (POMDP;~\cite{Astrom1965-ym}) is defined by the tuple \( (\mathcal{S}, \mathcal{A}, \mathcal{O}, T, O, R) \), where $\mathcal{S}$, $\mathcal{A}$, and $\mathcal{O}$ are the state, action, and observation spaces, and \( T: \mathcal{S} \times \mathcal{A} \to \Delta(\mathcal{S}) \),  \( O: \mathcal{S} \times \mathcal{A} \to \Delta(\mathcal{O}) \), and \( R: \mathcal{S} \times \mathcal{A} \to \mathbb{R} \) are the transition, observation and reward function respectively, see e.\,g.\ \cite{littman_thesis}.

The goal is to design a policy \( \pi: \mathcal{H} \to \mathcal{A} \) where \( \mathcal{H} \) is the set of histories of observations and actions, i.e. $\pi(a_t| o_t, a_{t-1}, o_{t-1}, ...)$, to maximize the expected reward:
\[
\pi^* = \arg\max_\pi \mathbb{E}\left[ \sum_{t=0}^\infty \gamma^t r(s_t, a_t) \middle | \pi \right]
\]
with \( \gamma \in [0, 1)\) being a discount factor.
As opposed to the fully observable case, the decision maker cannot reliably determine their own true state, $s$, based on the current observation alone, and a Markovian optimal policy does not exist in general. 
However, the agent can summarize the history by computing a belief state \( b \in \mathcal{B} = \Delta(S) \).
Ideally, it would compute the Bayesian update
\[
b(s'|a, o) \propto O(o|s', a) \int_{s \in S} T(s'|s, a) b(s),
\]
however, this is usually intractable to compute, and e.g. variational approximations are used in practice.

Since the Bayesian belief state serves as a sufficient statistic of the history, optimal policies for the original POMDP can be determined by solving an equivalent continuous-space MDP, known as a \textit{belief MDP} \( ( \mathcal{B}, \mathcal{A}, T_b, r_b ) \), where the new transition \( T_b \) and reward functions \( r_b \) are defined over $\mathcal{B}\times\mathcal{A}$. 
Using belief MDPs, several theoretical results about MDPs can be extended to POMDPs, such as the existence of optimal deterministic policies. 

In this continuous MDP, the goal is to maximize the cumulative reward by identifying a policy that uses the current belief state as input. 
Formally, we seek a policy \( \pi^* \) that satisfies 

\[
\pi^* = \arg\max_{\pi} \mathbb{E} \left[ \sum_{t=0}^{\infty} \gamma^t r_b(b_t, a_t) \right],
\]

where \( r_b(b_t, a_t) = \int_{s \in\mathcal{S}} b(s_t)r(s_t,a_t) \).

While POMDPs are very general, they can be difficult to map onto active sensing-type problems. 
Active SLAM introduces further specifications to the POMDP formalism in the form of belief-dependent rewards~\cite{Araya-Lopez_undated-qy}, and a structured state-space.
The state at time $t$ is usually assumed to be a combination of the robot's pose $x_t \in \mathcal{X}$ and a map $m_t \in \mathcal{M}$ that the robot builds up over time while interacting with the POMDP, resulting in a factored state space of the form $\mathcal{S}=\mathcal{X}\times\mathcal{M}$.

\end{document}